\documentclass[twocolumn]{article} 

\usepackage{multicol}  
\usepackage[letterpaper, top=0.6in, bottom=0.5in, inner=0.75in, outer=0.75in, headsep=0.3in, footskip=0.3in]{geometry}

\usepackage{xcolor}
\usepackage{listings}
\usepackage{caption}
\usepackage{subcaption}
\usepackage{enumitem}

\definecolor{codegray}{gray}{0.95}

\lstdefinestyle{promptblock}{
    backgroundcolor=\color{codegray},
    frame=tb, 
    framerule=0.5pt,
    rulecolor=\color{black!30},
    basicstyle=\small\ttfamily,
    keywordstyle=\color{blue!70!black},
    commentstyle=\color{green!60!black},
    stringstyle=\color{purple!80!black},
    breaklines=true,
    breakatwhitespace=true,
    tabsize=2,
    showstringspaces=false,
    morekeywords={system, user, assistant, <|im_start|>, <|im_end|>, , <tool_call>, <tool_response>},
    captionpos=b, 
    abovecaptionskip=\medskipamount,
}

\usepackage{titlesec}
\titleformat{\section}{\large\bfseries}{\thesection}{1em}{}
\titleformat{\subsection}{\normalsize\bfseries}{\thesubsection}{1em}{}

\def\huggingface{\raisebox{-1.5pt}{\includegraphics[height=1.05em]{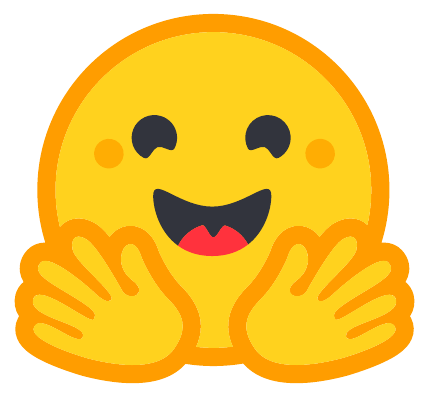}}}

\usepackage{tabularx} 

\usepackage{graphicx}         
\usepackage{amsmath, amssymb} 
\usepackage{hyperref}          
\usepackage{natbib}          
\usepackage{geometry}         
\usepackage{listings}
\usepackage{pgfplots}
\usepackage{amssymb} 
\usepackage{pifont}  
\usepgfplotslibrary{fillbetween}
\usepackage{algorithm}
\usepackage{algorithmic}
\usepackage{amsmath} 
\usepackage{appendix}
\usepackage{booktabs} 
\usepackage{caption}  
\usepackage{array}

\title{\textbf{Lucy: edgerunning agentic web search on mobile with machine generated task vectors}}

\author{
    Alan Dao (Gia Tuan Dao)\textsuperscript{1}, Dinh Bach Vu\textsuperscript{1}, Alex Nguyen\textsuperscript{1} , Norapat Buppodom\textsuperscript{1} \\[0.75em]
    Menlo Research \\ [0.5em]
    \texttt{alan@menlo.ai} \\ [0.5em]
\begin{tabular}{rl}
\huggingface & \url{https://huggingface.co/Menlo/Lucy} \\
\huggingface & \url{https://huggingface.co/Menlo/Lucy-128k} \\
\huggingface & \url{https://huggingface.co/Menlo/Lucy-gguf} \\
\huggingface & \url{https://huggingface.co/Menlo/Lucy-128k-gguf} \\
\end{tabular}
}

\date{\today}  

\usepackage{fancyhdr} 
\begin{document}

\pagestyle{fancy}
\fancyhf{}
\fancyhead[L]{\includegraphics[height=2.7em]{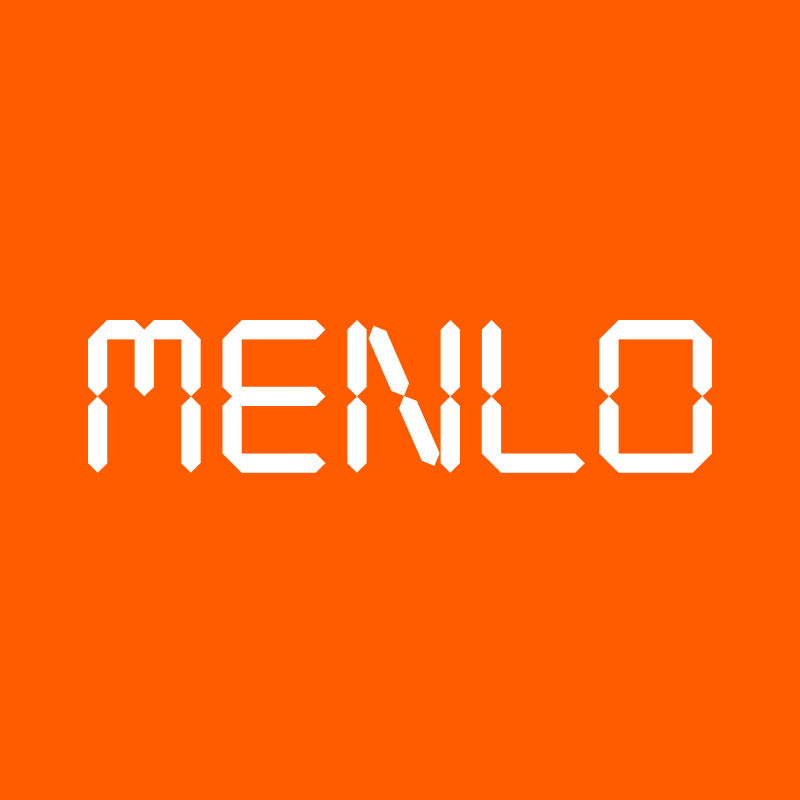}}
\fancyhead[R]{\today}
\fancyfoot[C]{\thepage}
\renewcommand{\headrulewidth}{0.5pt}
\fancypagestyle{plain}{\pagestyle{fancy}}

\maketitle
\begin{abstract}
Small language models (SLMs) are inherently limited in knowledge-intensive tasks due to their constrained capacity. While test-time computation offers a path to enhanced performance, most approaches treat reasoning as a fixed or heuristic process. In this work, we propose a new paradigm: viewing the model's internal reasoning, delimited by \texttt{<think>} and \texttt{</think>} tags, as a \textbf{dynamic task vector machine}. Rather than treating the content inside these tags as a mere trace of thought, we interpret the generation process itself as a mechanism through which the model \textbf{constructs and refines its own task vectors} on the fly. We developed a method to optimize this dynamic task vector machine through RLVR and successfully trained an agentic web-search model. We present \textbf{Lucy}, a 1.7B-parameter SLM that leverages this dynamic reasoning mechanism with MCP integration to achieve \textbf{78.3\%} accuracy on the SimpleQA benchmark, performing on par with much larger models such as DeepSeek-V3. This demonstrates that small models can rival large ones when equipped with structured, self-constructed task reasoning.
\noindent 
\end{abstract}

\section{Introduction}
\label{sec:introduction}

Large language models (LLMs) have demonstrated remarkable capabilities in natural language understanding and generation \citep{hendrycks2020measuring,clark2018think}. 
Yet, when faced with knowledge-intensive tasks requiring up-to-date information or multi-step reasoning, they often fall short due to inherent knowledge cutoffs and unstable inference dynamics \citep{jin2024long, wei2022chain}. 
A natural solution is to augment LLMs with external tools, particularly web search, enabling them to retrieve current facts and verify claims dynamically. 
This has given rise to the paradigm of \emph{agentic search}, where models alternate between reasoning and tool use to solve complex queries \citep{yao2023react, gao2023retrieval, schick2023toolformer}.

\begin{figure}
    \centering
    \includegraphics[width=1\linewidth]{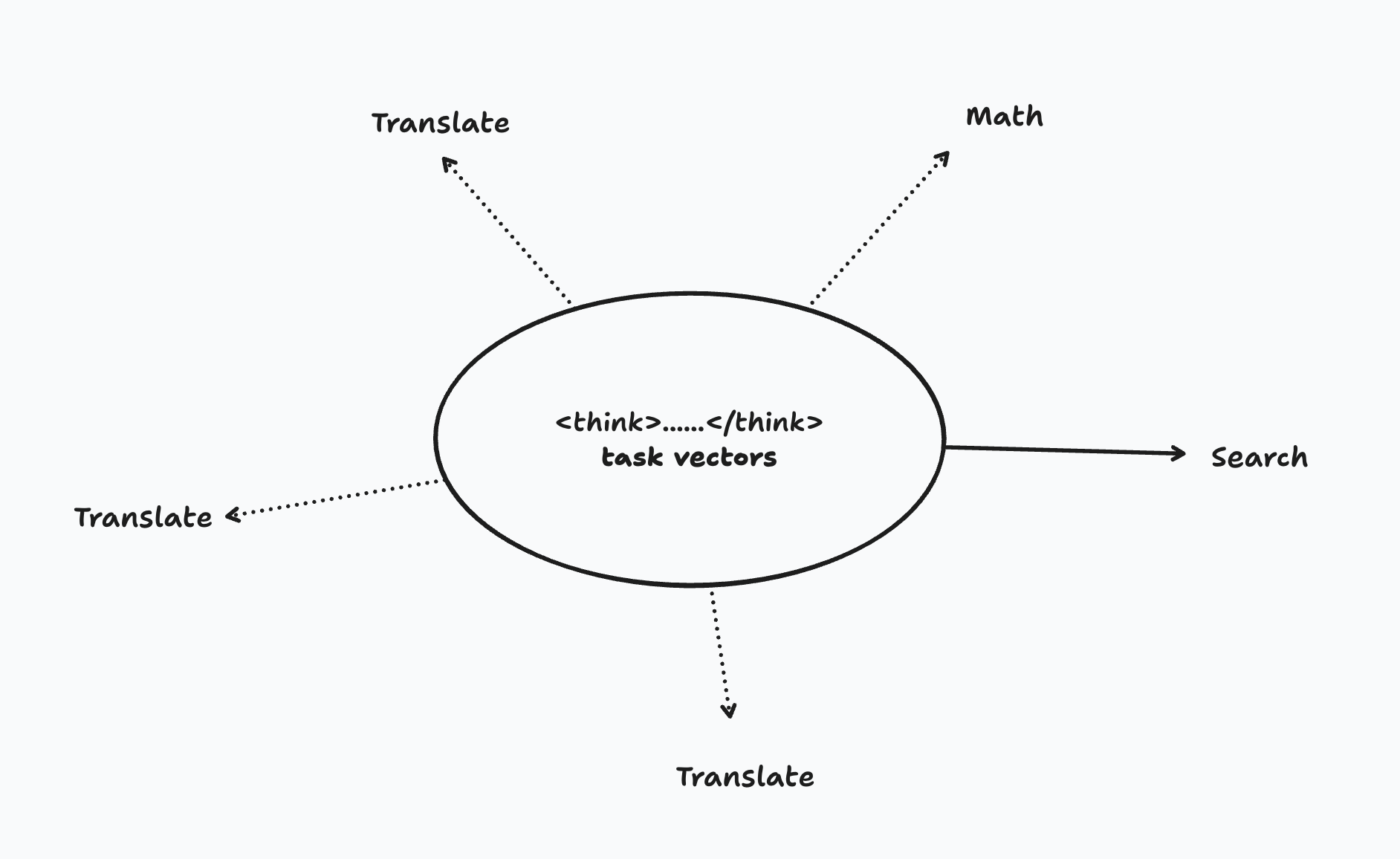}
    \caption{Optimiz generic think tag to be specific}
    \label{fig:concept-demo}
\end{figure}

However, a critical challenge remains: \textbf{how to stabilize the reasoning process} that guides search behavior. 
While models like ToolFormer \citep{schick2023toolformer} and ReAct \citep{yao2023react} show that LLMs can learn to call tools, their reasoning trajectories are often inconsistent, redundant, or divergent, especially over multiple turns. 

As emphasized by \citet{wu2025webdancerautonomousinformationseeking}, the central challenge in long-horizon information seeking does not stem from the ability to access search tools, but rather from the difficulty of \emph{constructing reliable reasoning trajectories}, that is, effectively decomposing complex tasks while preserving coherence with the overall objective. In addition, both \citet{wu2025webdancerautonomousinformationseeking,yin2025marcoo1v2wideningdistillation} and \citet{li2025larabenchmarkingretrievalaugmentedgeneration} observed that as reasoning chains grow longer, performance tends to degrade, in part because the growing volume of contextual information risks overwhelming the model’s context window.

In this work, we argue that the foundation for such reasoning already exists within the model. 
Modern LLMs are not just pattern marchers, they implicitly construct internal task representations during inference. 

This idea is supported by the concept of \emph{task vectors}, which have been shown to encode task-specific information in the latent space even before token generation begins \citep{hendel-etal-2023-context}. 
We extend this insight: rather than viewing the reasoning process as a passive trace, we treat the generation within \texttt{<think>} tags as a \textbf{dynamic task vector machine}, a self-modifying computational mechanism through which the model actively constructs, updates, and refines its task representation at test time.

The key challenge, then, is not to teach the model \emph{how} to search, but to \textbf{stabilize this internal task vector machine} so that the model can reliably generate coherent, goal-directed reasoning. 
We show that by structuring and optimizing the \texttt{<think>} process, through targeted training, architectural priors, and feedback mechanisms, we can significantly reduce noise and improve the consistency of reasoning, enabling effective multi-turn search and verification.

Our main contributions are:

- \textbf{A recipe for optimizing the task vector:} We introduce training strategies and architectural constraints that guide the model to generate more stable, self-consistent reasoning within the \texttt{<think>} loop, effectively enabling it to refine its own task representation during inference.
    
- \textbf{Theoretical and empirical analysis of task vector stability:} We analyze how a well-structured reasoning process reduces noise and improves credit assignment in multi-step tool use, showing that an optimized task vector machine leads to better alignment between reasoning steps and tool calls.
    
- \textbf{Evidence that small models can master agentic behavior:} We demonstrate that even a 1.7B-parameter model can achieve strong performance through reinforcement learning, challenging the prevailing assumption that only large models can support stable RL training and complex agentic reasoning.

- \textbf{Challenging the data bottleneck assumption:} We demonstrate that high-quality agentic behavior can emerge in a small 1.7B model trained on a limited, isolated dataset, proving that performance gains need not come from scaling data or model size, but from shifting the focus of optimization toward stabilizing the reasoning process.
\section{Related Work}
\label{sec:related_work}

Our work bridges three research areas critical for developing robust information-seeking agents, with particular emphasis on stabilizing the reasoning process through task vector optimization.

\subsection{Dynamic Reasoning in Language Models}
Recent work has revealed that LLMs construct internal task representations during inference, a phenomenon formalized through \emph{task vectors} \citep{hendel-etal-2023-context}. These latent representations govern the model's reasoning trajectory, particularly in agentic settings where models alternate between thinking and acting \citep{yao2023react}. While approaches like IRCoT \citep{trivedi2022interleaving} and ReAct demonstrate the value of explicit reasoning traces, they often suffer from inconsistency in multi-turn scenarios \citep{wu2025webdancerautonomousinformationseeking}. Our work extends these foundations by treating the reasoning process as a \emph{dynamic task vector machine} that actively maintains and refines its internal representation during search interactions.

\subsection{Search-Augmented Generation}
The integration of retrieval with LLMs has evolved through two paradigms: (1) Retrieval-Augmented Generation (RAG) systems \citep{lewis2020retrieval,xiong2025rag} that suffer from static retrieval limitations, and (2) dynamic tool-use approaches like Toolformer \citep{schick2023toolformer}. While RAG methods struggle with irrelevant context \citep{jin2024long}, tool-augmented models face challenges in maintaining reasoning coherence across multiple search steps \citep{jiang2023active}. Our approach differs by optimizing the \texttt{<think>} process as a stabilizing mechanism for the underlying task vector, enabling more consistent search behavior.

\subsection{Reinforcement Learning for Agentic Systems}
Building on RLHF foundations \citep{ouyang2022training}, recent advances have simplified policy optimization through methods like DPO \citep{rafailov2023direct} and GRPO \citep{shao2024deepseekmath}. However, these approaches typically optimize for final outcomes rather than reasoning stability. Works like LeRet \citep{hsu2024grounding} and R1-Searcher \citep{song2025r1} demonstrate RL's potential for search behavior, but none explicitly address the task vector dynamics crucial for coherent multi-step reasoning. Our framework introduces novel rewards that directly optimize the consistency and information density of the reasoning process itself.

\section{Methodology}
\label{sec:methodology}

Our methodology is founded upon the architectural principles of the Jan-Nano project~\cite{dao2025jan}, but diverges significantly in its approach to the model's reasoning process. We employ a multi-stage reinforcement learning system that circumvents the need for supervised fine-tuning (SFT) and integrates a local Retrieval-Augmented Generation (RAG) server for real-time information retrieval. However, where Jan-Nano offers a "non-thinking" approach by directly generating answers, our work focuses on retaining and optimizing the model's chain-of-thought (CoT) reasoning. A primary challenge in language models that utilize chain-of-thought reasoning is the tendency for "overthinking," where the model generates excessively verbose or redundant reasoning steps, leading to increased computational costs and latency. We introduce a novel two-stage training process with carefully designed reward functions to foster efficient thinking and mitigate the common issue of model "overthinking".
\subsection{Composite Reward}

Our reinforcement learning framework employs a multi-component reward function that combines several specialized reward terms to comprehensively shape agent behavior. This reward structure consists of two main categories:
\subsubsection{Foundational Reward Components}
The primary training signal comprises core reward terms that enforce correctness, structural compliance, and proper tool usage. These components build upon the reward design from the Jan-Nano baseline, adapted to support explicit reasoning chain preservation:
\begin{itemize}
    \item \textbf{Correctness Reward ($r_{\text{correct}}$):} This is the primary objective signal. For the QA task, it is a binary reward ($1.0$ or $0.0$) determined by a substring match between the model's generated answer and the ground truth. This accommodates minor variations in phrasing while ensuring semantic accuracy.
    
    \item \textbf{XML Validity Reward ($r_{\text{xml}}$):} To ensure structural integrity, this function checks for well-formed XML tags (\texttt{<think>}, \texttt{<tool\_call>}, \texttt{<answer>}). It penalizes unbalanced tags or logically inconsistent generations (e.g., issuing a \texttt{<tool\_call>} and an \texttt{<answer>} in the same turn) by returning 0. For valid structures, it returns a normalized score encouraging a complete reasoning cycle:
    \begin{equation}
    r_{\text{xml}} = \frac{N_{\text{answer}} \times (N_{\text{think}} + N_{\text{tool}})}{N_{\text{turn}}}
    \end{equation}
    where $N$ is the count of respective tags or turns.

    \item \textbf{Format Adherence Reward ($r_{\text{format}}$):} This function assesses the overall compliance of the model's output with the predefined response schema. It provides a continuous score reflecting the degree of correctness, with a specific value indicating a perfectly formatted response.

    \item \textbf{Tool Execution Reward ($r_{\text{tool}}$):} A reward is provided for successfully executing a called tool, determined by a valid, non-error response from the environment.
\end{itemize}
\subsubsection{Behavior-Centric Reward Functions}

In addition to standard correctness and formatting objectives, we incorporate auxiliary reward signals that incentivize efficient search behavior and purposeful reasoning:

\paragraph{Visit/Search Ratio.}
To promote judicious use of the search tool, we apply a reward function based on the ratio between document visits and issued search queries. If the agent performs more searches than visits, it receives a fixed penalty of $-0.5$. Otherwise, the reward is computed as:
\[
r_{\text{visit/search}} = \left( \frac{\text{visit\_search\_ratio} - 1}{4} \right)^{0.25}
\]
This reward gradually increases as the agent favors visiting over excessive querying, encouraging focused exploration strategies.

\paragraph{Efficient Thinking.}
To promote concise yet effective reasoning, we introduce a reward signal based on the length of the model's internal reasoning span, delimited by \texttt{<think>} and \texttt{</think>} tags. Let \( x \) denote the number of tokens within this span.

The reward is defined as:
\[
r_{\text{think}}(x) = \operatorname{SN}(x;\, \mu = 35,\, \sigma = 150,\, \alpha = -5)
\]
where \( \operatorname{SN}(\cdot) \) is the skew-normal probability density function. The distribution is centered at 35 tokens with a negative skew, encouraging informative but succinct reasoning. Excessively long reasoning sequences are penalized more than shorter ones, avoiding unnecessary verbosity.

The skew-normal form was chosen heuristically. In preliminary experiments with \textsc{Jan-Nano}, we observed a recurring issue of overthinking, where models produced unnecessarily long chains of reasoning that degraded task performance, especially around tool use. These overextensions often involved questioning whether to use tools or attempting to answer without them, leading to degraded output.

By shaping the reward to discourage such behavior, we align the model’s behavior toward efficient, context-appropriate reasoning. While the hyperparameters were set empirically, they reflect a broader assumption: different tool contexts benefit from different reasoning lengths, but moderate spans are generally preferable.
\subsection{Two-Stage Reinforcement Learning Framework}

Our training is divided into two distinct stages, each with a specific objective. This two-stage approach allows the model to first learn the foundational skills of correctness and tool use before polishing its formatting and accuracy.

\subsubsection*{Stage 1: Foundational Training for Correctness and Tool Proficiency}

The initial stage is designed to teach the model the foundational skills of correctness and effective tool use while gently shaping its reasoning behavior. The total reward, $R_1$, is formulated to make correctness a prerequisite for receiving credit for other desirable behaviors.

Let $r_{\text{correct}}$ be the correctness reward. We define an auxiliary behavioral score, $b$, as a weighted sum of the secondary rewards:
\begin{equation}
b = w_{\text{tool}}r_{\text{tool}} + w_{\text{format}}r_{\text{format}} + w_{\text{think}}r_{\text{think}} + w_{\text{xml}}r_{\text{xml}} + w_{\text{v/s}}r_{\text{v/s}}
\end{equation}

The weights are set to $[w_{\text{tool}}=0.2, w_{\text{format}}=0.2, w_{\text{think}}=0.1, w_{\text{xml}}=0.1, w_{\text{v/s}}=3.0]$, with the high weight on $r_{\text{v/s}}$ strongly prioritizing efficient search.

The final reward for Stage 1 is then:
\begin{equation}
R_1 = r_{\text{correct}} \times \log(1.001 + r_{\text{correct}} \times b)
\end{equation}

A critical condition is applied: if the XML structure is invalid ($r_{\text{xml}}=0$), the behavioral score $b$ is assigned a fixed penalty of $-0.5$.

This formulation has two key properties. First, if the answer is incorrect ($r_{\text{correct}}=0$), the total reward $R_1$ is zero, ensuring that the model cannot accumulate rewards for good behavior that leads to a wrong answer. Second, the logarithmic function provides a non-linear, diminishing return for the behavioral score $b$. This encourages the model to improve its formatting, thinking, and tool use, but prevents these secondary objectives from overshadowing the primary goal of correctness.

\subsubsection*{Stage 2: Fine-tuning for Format Adherence and Maximizing Accuracy}

After reaching convergence in Stage 1, the model enters a refinement phase aimed at ensuring compliance with output constraints and optimizing performance. The reward function shifts from a continuous gradient-based signal to a binary threshold mechanism that either accepts or rejects outputs based on specification adherence.

The total reward for Stage 2, $R_2$, is defined as a product of binary gates:
\begin{equation}
R_2 = r_{\text{correct}} \times g_{\text{format}} \times g_{\text{xml}}
\end{equation}

where:
\begin{itemize}
    \item $r_{\text{correct}}$ remains the binary correctness reward.
    \item $g_{\text{format}}$ is a binary gate: $g_{\text{format}} = 1$ if the format is perfect ($r_{\text{format}}$ meets a specific threshold), and $0$ otherwise.
    \item $g_{\text{xml}}$ is a binary gate: $g_{\text{xml}} = 1$ if the XML structure is valid ($r_{\text{xml}} > 0$), and $0$ otherwise.
\end{itemize}

The transition to a binary reward structure in Stage 2 enforces strict output constraints on the model. To receive positive reward, the model must simultaneously satisfy three requirements: factual correctness, valid XML syntax, and exact format compliance. This rigid criterion removes reward signal ambiguity and drives the model toward a consistent, reliable output distribution, enhancing its utility for production deployment.
\subsection{Data}

We use the same dataset utilized in training of Jan Nano~\cite{dao2025jan}, which is an adaptation of the MuSiQue-Ans dataset~\cite{trivedi2022musiquemultihopquestionssinglehop}. Our final dataset consisted of 10,325 question-answer pairs.
\section{Experiments and Results}
\label{sec:experiments}

\subsection{Experiments}
We evaluated our model on the SimpleQA dataset, following the evaluation protocol established in the Jan-Nano project. To assess the models' practical tool-using abilities, we integrated our evaluation code with a MCP(Model Context Protocol) server, which provides a standardized interface for web search and scraping tools.

Lucy achieves 78.3\% accuracy despite its compact 1.7B parameter count. This represents a substantial 19.1 percentage point improvement over the 4B parameter baseline and matches the performance of the significantly larger DeepSeek-67B model (78.2\%). Our key finding demonstrates exceptional parameter efficiency, validating our training approach: by preserving and optimizing chain-of-thought reasoning through behavior-targeted rewards, we successfully elicit complex reasoning and tool usage from a compact architecture.
Lucy's competitive performance relative to 4B-parameter Jan-Nano models indicates that our reward design effectively transfers principled search behaviors to smaller architectures, challenging assumptions that high-performance agentic capabilities require large-scale models. This combination of reduced model size and maintained accuracy enables practical deployment of capable autonomous agents on mobile and edge devices, providing advantages in latency, privacy, and accessibility.

\begin{table}[H]
    \centering
    \begin{tabular}{p{3.7cm}cc}
    \toprule
    \textbf{Model} & \textbf{SimpleQA} & \textbf{Parameters} \\
    \midrule
    OpenAI o1 & 42.6\% & Unknown \\
    Grok 3 & 44.6\% & Unknown \\
    o3 & 49.4\% & Unknown \\
    Claude-3.7-Sonnet & 50.0\% & Unknown \\
    Gemini-2.5 Pro & 52.9\% & Unknown \\
    ChatGPT-4.5 & 62.5\% & Unknown \\
    \midrule
    \multicolumn{3}{l}{\textbf{With MCP:}} \\
    DeepSeek-671B (OpenRouter) & 78.2\% & 671B \\
    \textbf{Lucy (think mode)} & \textbf{78.3}\% & 1.7B \\
    Jan-nano & 80.7\% & 4B \\
    Jan-nano-128k & 83.2\% & 4B \\
    \bottomrule
    \end{tabular}
    \caption{SimpleQA benchmark~\cite{wei2024measuringshortformfactualitylarge} results.}
    \label{tab:benchmarks}
\end{table}


\section{Discussion}

\subsection{Emergent Skipping of Redundant Thinking}
During optimization of Lucy's reasoning process, we observed an intriguing adaptation in \textit{one representative scenario} involving multi-step search tasks. When penalizing excessively long reasoning spans, the model learned to strategically omit thinking steps during low-decision operations—a behavior distinct from our initial expectation of uniformly shorter reasoning traces.

\subsubsection*{Illustrative Example}
Consider a 5-step research task where the model alternates between searching and reading:

\begin{itemize}
    \item \textbf{Baseline Behavior:} Full deliberation at each step:
    \begin{enumerate}[label=(\roman*)]
        \item \texttt{<think>} Search about topic \texttt{</think>}
        \item \texttt{<think>} Read result \texttt{</think>}
        \item \texttt{<think>} Read result \texttt{</think>}
        \item \texttt{<think>} Search details \texttt{</think>}
    \end{enumerate}

    \item \textbf{Optimized Behavior:} Thinking suppressed for reading:
    \begin{enumerate}[label=(\roman*)]
        \item \texttt{<think>} Search about topic \texttt{</think>}
        \item Read result \quad \textit{(no thinking tags)}
        \item Read result \quad \textit{(no thinking tags)}
        \item \texttt{<think>} Search details \texttt{</think>}
    \end{enumerate}
\end{itemize}

This \textit{specific case} suggests Lucy may dynamically allocate reasoning capacity based on action predictability. The model reserved thinking for high-uncertainty operations (e.g., query formulation) while bypassing deliberation for deterministic actions (e.g., processing retrieved content). Notably, this behavior emerged organically from the efficiency reward rather than explicit training. While this pattern improved latency by 17.8$\times$ in the observed scenario, its generalizability to other task types requires further validation. The finding highlights the need for context-aware efficiency incentives rather than universal reasoning compression.

\subsection{The Illusion of Test-Time Compute}
The common assumption that "more thinking improves performance" proves unreliable for small models. Our benchmarking revealed failure cases where Lucy couldn't formulate correct queries not due to missing internet data, but because it lacked fundamental knowledge about the queried entities—demonstrating that test-time compute can't compensate for missing conceptual grounding. This aligns with Anthropic's findings on inverse test-time scaling \cite{gema2025inversescalingtesttimecompute}, where extended reasoning provided diminishing returns.

Larger models like Jan-nano \cite{dao2025jan} surpass this limitation by self-correcting during research, while Lucy (1.7B) often circulates incorrect assumptions. The training process barely kept pace with Jan-nano's baseline, suggesting small models hit fundamental knowledge barriers no amount of reasoning can overcome.
\section{Conclusion}
\label{sec:conclusion}

Lucy demonstrates that small language models, when equipped with structured reasoning and reinforcement learning techniques, can rival much larger systems on complex knowledge-intensive tasks. By reframing the model’s internal reasoning process as a \textit{dynamic task vector machine}, we enable Lucy to iteratively construct and refine its goals during inference. Our use of \texttt{<think>} tags as both a scaffolding and optimization target---combined with behavior-centric rewards and a structured XML dialogue format---results in a highly efficient, agentic web-search model.

Despite its modest 1.7B parameter size, Lucy achieves 78.3\% accuracy on the SimpleQA benchmark under MCP settings, matching models several hundred times larger. These results challenge conventional assumptions around scale, data requirements, and test-time reasoning. We also uncover intriguing emergent behaviors, such as dynamic skipping of redundant thought, and reveal limits to ``thinking harder'' without grounding in knowledge.

Ultimately, Lucy suggests a new direction for small, capable models: not through brute-force parameter scaling, but through \textit{training models to think better---not longer}. We hope this work inspires further exploration into lightweight, agentic systems optimized for real-world interaction and tool use.



\bibliographystyle{plainnat} 
\bibliography{main} 


\clearpage
\appendix
\onecolumn

\appendix
\section{Appendix: Qwen3-4B Thinking Tag Analysis on Jan-Nano dataset}
\label{app:qwen3-4b think tag}

We hypothesized that content inside recent reasoning model's \texttt{<think></think>} tag is "task vector", a concept discussed by \cite{hendel2023context}. If this hypothesis is correct, moving contextual information such as tool calls and their response into or out of the tag should affect the performance of the model's performance.

To test this, we analyzed the effect on different prompt templates on Qwen3-4B with SimpleQA benchmark. The procedure was:

\begin{enumerate}
\item We sampled 500 questions from the evaluation logs of the Jan-Nano-128k run on the \textsc{SimpleQA} benchmark. This provided a consistent set of contextual information (tool calls and tool responses) for each question.
\item We prompted Qwen3-4B to generate a final answer using this same context, but with five different prompt strategies. These prompts is different on how tool calls and responses were placed relative to the \texttt{<think>} tags.
\item We evaluated the accuracy of the model's generated answers against the ground-truth short answers, using the same evaluation pipeline from the Jan-Nano benchmark.
\end{enumerate}

\noindent
The five templates evaluated are reproduced in Table~\ref{tab:dryrun_templates}.  Scores in bold denote accuracy.

\begin{table}[htbp]
\centering
\small
\begin{tabularx}{\textwidth}{X c c c}

\toprule
\textbf{Template} & \textbf{Model} & \textbf{ Accuracy} & \textbf{Observation} \\
\midrule

1. Baseline Prompt \ref{fig:template1} & Jan-Nano-128K          & \textbf{81.2\%}  & - \\
\midrule
1. Baseline Prompt \ref{fig:template1} & Qwen3-4B          & \textbf{81.6\%}  & - \\
2. Think Prompt \ref{fig:template2} & Qwen3-4B & \textbf{82.2\%}  & - \\
3. Tools and responses inside \texttt{<think>} \ref{fig:template3} & Qwen3-4B & \textbf{77.4\%}  &  Model tries to call tools \\
4. Tools outside, responses inside \texttt{<think>}  \ref{fig:template4} & Qwen3-4B & \textbf{73.8\%}  & Model returns empty string \\
5. Sequenced \texttt{<think>} blocks \ref{fig:template5} & Qwen3-4B & \textbf{63.0\%}  & Model returns empty string \\
\bottomrule
\end{tabularx}
\caption{Dry-run results on the eval logs of \textsc{Jan-Nano-128K} on \textsc{SimpleQA}(500 samples).}
\label{tab:dryrun_templates}
\end{table}

\begin{figure}[htbp]
    \centering
    \begin{lstlisting}[style=promptblock, caption={Baseline Prompt. Tool interactions follow standard user/assistant turns.}, label={fig:template1}]
<|im_start|>system {system_prompt}<|im_end|>

<|im_start|>user
What episode of the TV show "Sister, Sister" did Chip Fields-Hurd<|im_end|>

<|im_start|>assistant
<tool_call>{tool_call_1}</tool_call>

<|im_start|>user
<tool_response>{tool_response_1}</tool_response>

<|im_start|>assistant
<tool_call>{tool_call_2}</tool_call>

<|im_start|>user
<tool_response>{tool_response_2}</tool_response>

<|im_start|>assistant
    \end{lstlisting}
\end{figure}

\begin{figure}[htbp]
    \begin{lstlisting}[style=promptblock, caption={Think Prompt. All tool interactions are wrapped in a single \texttt{<think>} block, while leaving think tag opens, allowing model to continue its thinking}, label={fig:template2}]
<|im_start|>system {system_prompt}<|im_end|>

<|im_start|>user
What episode of the TV show "Sister, Sister" did Chip Fields-Hurd<|im_end|>

<|im_start|>assistant

<think>

<tool_call>{tool_call_1}</tool_call>
<tool_response>{tool_response_1}</tool_response>

<tool_call>{tool_call_2}</tool_call>
<tool_response>{tool_response_2}</tool_response>

Reasoning:
    \end{lstlisting}
\end{figure}
    
\begin{figure}[htbp]
    \begin{lstlisting}[style=promptblock, caption={Template 3: All tool calls and responses inside \texttt{<think></think>} tag. The model must reason and produce the final answer from within the tag.}, label={fig:template3}]
<|im_start|>system {system_prompt}<|im_end|>

<|im_start|>user
What episode of the TV show "Sister, Sister" did Chip Fields-Hurd<|im_end|>

<|im_start|>assistant

<think>

<tool_call>{tool_call_1}</tool_call>
<tool_response>{tool_response_1}</tool_response>

<tool_call>{tool_call_2}</tool_call>
<tool_response>{tool_response_2}</tool_response>

End of tools call.\n</think>
    \end{lstlisting}
\end{figure}
    
\begin{figure}[htbp]
    \begin{lstlisting}[style=promptblock, caption={Template 4: Tools outside, responses inside \texttt{}. Tool calls are outside \texttt{<think></think>}, but tool responses are grouped inside it. This break order of the tool call/response.}, label={fig:template4}]
<|im_start|>system {system_prompt}<|im_end|>

<|im_start|>user
What episode of the TV show "Sister, Sister" did Chip Fields-Hurd<|im_end|>

<|im_start|>assistant

<think>
</think>

<tool_call>{tool_call_1}</tool_call>
<tool_call>{tool_call_2}</tool_call>

<think>

<tool_response>{tool_response_1}</tool_response>
<tool_response>{tool_response_2}</tool_response>

Now let me think about this information and provide an answer:
    \end{lstlisting}
\end{figure}

\begin{figure}[htbp]
    \begin{lstlisting}[style=promptblock, caption={Template 5: Sequenced \texttt{} blocks. Each tool call is followed by a \texttt{<think></think>} block containing its response.}, label={fig:template5}]
<|im_start|>system {system_prompt}<|im_end|>

<|im_start|>user
What episode of the TV show "Sister, Sister" did Chip Fields-Hurd<|im_end|>

<|im_start|>assistant

<think>
</think>

<tool_call>{tool_call_1}</tool_call>

<think>
<tool_response>{tool_response_1}</tool_response>
</think>

<tool_call>{tool_call_1}</tool_call>

<think>
<tool_response>{tool_response_2}</tool_response>

Now let me think about this information and provide an answer:
    \end{lstlisting}
\end{figure}

\newpage

\section{Appendix: Case Study on Task Vector Steering for AIME-2024}
\label{sec:appendix_case_study}

We sampled mathematical reasoning problems from AIME 2024 to evaluate the effectiveness of task vector steering. We compare a baseline zero-shot prompt against a modified prompt where a task vector, derived from the model's own chain-of-thought (CoT) processing for the *same problem*, is injected into its hidden states. The goal is to induce a "reasoning" behavior without the overhead of a full CoT prompt.

The experiment was conducted using the Qwen-3 1.7B model. A "reasoning" task vector was created by taking the layer 23 output of the last token with a CoT-prompted run. This vector was then added as layer 23 output of the last token during a standard zero-shot inference pass. The results are summarized in Table~\ref{tab:appendix_results}.

\begin{table}[h!]
\centering
\caption{Model Performance Comparison on AIME-2024 Problems}
\label{tab:appendix_results}
\begin{tabular}{@{}lcccc@{}}
\toprule
\textbf{Problem ID} & \textbf{Ground Truth} & \textbf{With Think (CoT)} & \textbf{Baseline (No Think, T=0.6)} & \textbf{Baseline + Task Vector} \\
\midrule
2024-I-1 & 204 & 204 (\checkmark) & 204 (\text{\sffamily X} initially) & 204 (\checkmark) \\
2024-I-4 & 116 & 116 (\checkmark) & 172 (\text{\sffamily X}) & 116 (\checkmark) \\
\bottomrule
\end{tabular}
\textit{{Note}: For problem I-1, the baseline model required multiple attempts to succeed; the table reflects its initial failure. The checkmark (\checkmark) indicates the method produced the correct answer, while the (\text{\sffamily X}) denotes a failure.}
\end{table}

As shown, The model steered by the task vector produced the correct answer more reliably and concisely than the verbose CoT method. This suggests the vector successfully guided the model toward a more robust computational process. 

While still a limited sample, these two case studies support hypothesis that task vector steering can be a powerful technique. It shows promise for improving model reliability on complex tasks in an efficient manner.

\newpage
\null
\thispagestyle{empty}
\newpage
\begin{center}
  \resizebox{\textwidth}{!}{%
    \begin{minipage}{1.95\textwidth} 
\begin{lstlisting}
#%%%%%%%%%%%%%%%%%%%%%%%%%%%%%%%%%%%%%%%%%%%%%%%%%%%%%%%%%%%%%%%%%%%%%%%%%%%%%%%%%%%%%%%%%%%%%%%%%%%%%%%%%%%%%%%%%%%%%%%%%%%%%%%%%%%%%%%%%%%%%%%%%%%%%%%%%%%%%
#%%%%%%%%%%%%%%%%%%%%%%%%%%%%%%%%%%%%%%%%%%%%%%%%%%%%%%%%%%%%%%%%%%%%%%%%%%%%%%%%%%%%%%%%%%%%%%%%%%%%%%%%%%%%%%%%%%%%%%%%%%%%%%%%%%%%%%%%%%%%%%%%%%%%%%%%%%%%%
#%%%%%%%%%%%%%%%%%%%%%%%%%%%%%%%%%%%%%%%%%%%%%%%%%%%%%%%%%%%%%%%%%%%%%%%%%%%%%%%%%%%%%%%%%%%%%%%%%%%%%%%%%%%%%%%%%%%%%%%%%%%%%%##****++++******###%%%%%%%%%%%%
#%%%%%%%%%%%%%%%%%%%%%%%%%%%%%%%%%%%%%%%%%%%%%%%%%%%%%%%%%%%%%%%%%%%%%%%%%%%%%%%%%%%%%%%%%%%%%%%%%%%%%%%%%%%%%%%%%%%%%%#*+++++++++++++++++++++++++++***++**#%%
#%%%%%%%%%%%%%%%%%%%%%%%%%%%%%%%%%%%%%%%%%%%%%%%%%%%%%%%%%%%%%%%%%%%%%%%%%%%%%%%%%%%%%%%%%%%%%%%%%%%%%%%%%%%%%%%%#+++++++++++++++++++++++++++++++++++++#*+++++
#%%%%%%%%%%%%%%%%%%%%%%%%%%%%%%%%%%%%%%%%%%%%%%%%%%%%%%%%%%%%%%%%%%%%%%%%%%%%%%%%%%%%%%%%%%%%%%%%%%%%%%%%%%#*++++++++++=++++++++++++++++++*++++++++++++*+**+++
#%%%%%%%%%%%%%%%%%%%%%%%%%%%%%%%%%%%%%%%%%%%%%%%%%%%%%%%%%%%%%%%%%%%%%%%%%%%%%%%%%%%%%%%%%%%%%%%%%%%%%%#+++++++++==+++++++++++++++++++++++*+****++++++***+***+
#%%%%%%%%%%%%%%%%%%%%%%%%%%%%%%%%%%%%%%%%%%%%%%%%%%%%%%%%%%%%%%%%%%%%%%%%%%%%%%%%%%%%%%%%%%%%%%%%%%#*+++++++++===-===+++++++++++++++++++*********+******+++***
#%%%%%%%%%%%%%%%%%%%%%%%%%%%%%%%%%%%%%%%%%%%%%%%%%%%%%%%%%%%%%%%%%%%%%%%%%%%%%%%%%%%%%%%%%%%%%%#*++++++++==---==+++++++++++++++++++++++++*********++++********
#%%%%%%%%%%%%%%%%%%%%%%%%%%%%%%%%%%%%%%%%%%%%%%%%%%%%%%%%%%%%%%%%%%%%%%%%%%%%%%%%%%%%%%%%%%%#*++++=====--:-=++++++++++++++=====-=+++**###*******++++++*****###
#%%%%%%%%%%%%%%%%%%%%%%%%%%%%%%%%%%%%%%%%%%%%%%%%%%%%%%%%%%%%%%%%%%%%%%%%%%%%%%%%%%%%%%%%#*+========-:::-=++++++++==--:::::-=+*##################*+++++**##***
#%%%%%%%%%%%%%%%%%%%%%%%%%%%%%%%%%%%%%%%%%%%%%%%%%%%%%%%%%%%%%%%%%%%%%%%%%%%%%%%%%%%%%%*+========-:::-===++===--::::::::=**#*******+***++++***#####+++*****+*+
#%%%%%%%%%%%%%%%%%%%%%%%%%%%%%%%%%%%%%%%%%%%%%%%%%%%%%%%%%%%%%%%%%%%%%%%%%%%%%%%%%%%*+========-::::-=++++=-:::::::::-**##**+*++++++++++++++++****##*+*++****++
#%%%%%%%%%%%%%%%%%%%%%%%%%%%%%%%%%%%%%%%%%%%%%%%%%%%%%%%%%%%%%%%%%%%%%%%%%%%%%%#%#+=========-::::-=====-::::::::::*#*+++++++++++++=--:--==+++***++*++++*++++*+
#%%%%%%%%%%%%%%%%%%%%%%%%%%%%%%%%%%%%%%%%%%%%%%%%%%%%%%%%%%%%%%%%%%%%%%%%%%%%#%#==========:::::-====-::::::::::=#*+++++++++=-:::::::=+=+++++++++****+++**+++++
#%%%%%%%%%%%%%%%%%%%%%%%%%%%%%%%%%%%%%%%%%%%%%%%%%%%%%%%%%%%%%%%%%%%%%%%%%%#%#=======+=-:::::-===-:.:::::::::**+==++===:::::::::-=====++++++########*+++++++++
#%%%%%%%%%%%%%%%%%%%%%%%%%%%%%%%%%%%%%%%%%%%%%%%%%%%%%%%%%%%%%%%%%%%%%%%%#%*=========::::::-===-::::::::::-**+=====::::::::::-=======++++*###**++++**+++++*###
#%%%%%%%%%%%%%%%%%%%%%%%%%%%%%%%%%%%%%%%%%%%%%%%%%%%%%%%%%%%%%%%%%%%%%%#%#+===-===-:::::::-=-:::::::::::-*+====-::::::::::-=========++*##**++++++++++=++++++++
#%%%%%%%%%%%%%%%%%%%%%%%%%%%%%%%%%%%%%%%%%%%%%%%%%%%%%%%%%%%%%%%%%%%%#%#+=======-:::::::---:::::::::::-++===-:::::::::::-==========+*#**++++++++=+++==++++++++
#%%%%%%%%%%%%%%%%%%%%%%%%%%%%%%%%%%%%%%%%%%%%%%%%%%%%%%%%%%%%%%%%%%%%#+====--=-::::::::--::::::::::::++===-:::::::::::-==========+**+===++++++++++++==+++==+++
#%%%%%%%%%%%%%%%%%%%%%%%%%%%%%%%%%%%%%%%%%%%%%%%%%%%%%%%%%%%%%%%%###*+======-:::::::::-:::::::::::.++==-::::::::::::-==========+**=--======-----=++=--+++====+
#%%%%%%%%%%%%%%%%%%%%%%%%%%%%%%%%%%%%%%%%%%%%%%%%%%%%%%%%%%%%%%%#%#+====-=-:::::::::-::::::::::::-*==-:::::::::::::-=========+*+-:-=====-::::::-===-:=++======
#%%%%%%%%%%%%%%%%%%%%%%%%%%%%%%%%%%%%%%%%%%%%%%%%%%%%%%%%%%%%%#%%+===--=-:::::::::-:::::::::::::++==:::.::.::::::---==---=-+*=::--====:::::::::-==-::=========
#%%%%%%%%%%%%%%%%%%%%%%%%%%%%%%%%%%%%%%%%%%%%%%%%%%%%%%%%%%%%#%#+==----::::::::::-:::...::.:::-+==::.::::::::...----::.===*=:::--=--:::::::::::===:::====--===
#%%%%%%%%%%%%%%%%%%%%%%%%%%%%%%%%%%%%%%%%%%%%%%%%%%%%%%%%%%##%*====-=-:::.....:::::::.:..:.:.-+=::.:::..:::.:.:---::::====:::-:---::::::::::::-==:::-====:-===
#%%%%%%%%%%%%%%%%%%%%%%%%%%%%%%%%%%%%%%%%%%%%%%%%%%%%%%%%%###=====---:.......:-.::::::::..::*+-:::.::::::.:::---::::--=+:::::---::::::::::::::==::::-===::--==
#%%%%%%%%%%%%%%%%%%%%%%%%%%%%%%%%%%%%%%%%%%%%%%%%%%%%%%%##%#====---:........:::::::::.:::::+=::....::.....::--:..::-==::::::--:::::::::::::::-=:::::===-:::-==
#%%%%%%%%%%%%%%%%%%%%%%%%%%%%%%%%%%%%%%%%%%%%%%%%%%%%%%##%*===----:.:......:::::::::.:.:.-=-:...:::::..::::-:...::---::::::-:::::::::::::::::-:::::-==-::::-==
#%%%%%%%%%%%%%%%%%%%%%%%%%%%%%%%%%%%%%%%%%%%%%%%%%%%%###%+==-----.........::..:.::.:::..-=:::::.:::::::..::::...:-=-:::::::::.::::::::::::::--:::::-=-:::::-==
#%%%%%%%%%%%%%%%%%%%%%%%%%%%%%%%%%%%%%%%%%%%%%%%%%%%####+=-----::.:......::..:.....:...:=:.....:::::::::::..:..:--::::::::::::.::::::::::::--::::::=-::::::-==
#%%%%%%%%%%%%%%%%%%%%%%%%%%%%%%%%%%%%%%%%%%%%%%%%%%####=-------:........-:.....:......:=.........::.::::......-=:::::::::::::::::::.:::::::-::::::--:::::::-==
#%%%%%%%%%%%%%%%%%%%%%%%%%%%%%%%%%%%%%%%%%%%%%%%%%##%#=-------:..:.....-.............:=:......::::::::::.:.::-:::.:.:::.:.::.......:::::::::::::::=-:::::::-==
#%%%%%%%%%%%%%%%%%%%%%%%%%%%%%%%%%%%%%%%%%%%%%%%%##%#=-------::.:.....:.............:+::........:..::::.::.:::::.::...::::..:.:..:.::::::::::::::--::::::::-==
#%%%%%%%%%%%%%%%%%%%%%%%%%%%%%%%%%%%%%%%%%%%%%%%##%#==-----=.::..:...:.............:=::.....::..::.:.:::::::::..........::::.:::.:.:::::::.::::::-:::::::::--=
#%%%%%%%%%%%%%%%%%%%%%%%%%%%%%%%%%%%%%%%%%%%%%%###*==-----=.:...:...:.... ........:-::........::.:.....:..=..........:::...:.:::::::::::::::::::-::::::::::-==
#%%%%%%%%%%%%%%%%%%%%%%%%%%%%%%%%%%%%%%%%%%%%%###*==------:.:.::...-.:...........:-::...................:-........:....::::..::::::::::::::::::-:::::::::::---
#%%%%%%%%%%%%%%%%%%%%%%%%%%%%%%%%%%%%%%%%%%%%###*==-=----::::::...-::............--....................::.::......::.:.:::..:::::::::::::::::::::::::::::::---
#%%%%%%%%%%%%%%%%%%%%%%%%%%%%%%%%%%%%%%%%%%###%*=----=---::..:..:-:::..:........:=:...................-:.:.........:..::::::::::::::::::::::::::::::::::::----
#%%%%%%%%%%%%%%%%%%%%%%%%%%%%%%%%%%%%%%%%%###%+==----===::..:..:-::::...:......:=.................:.:-:...............::.:::::::::-:::::::::::..::::::::.:---=
#%%%%%%%%%%%%%%%%%%%%%%%%%%%%%%%%%%%%%%%%%##%+==--=====-:.:...:-:...:......:..:=-.......::..........-................:..:::::.:::::::::::::.:...::::::...----=
#%%%%%%%%%%%%%%%%%%%%%%%%%%%%%%%%%%%%%%%####+==---====-:.:.:.:--:.::.::.:.....:+:.....:::.....::...-::....................:::::::::::::::..:....:::.....:-----
#%%%%%%%%%%%%%%%%%%%%%%%%%%%%%%%%%%%%%%####*=========--:::::.:-::::::...:.:::.+:.....:::...:.::::.-........................:::::..........-............:-----:
#%%%%%%%%%%%%%%%%%%%%%%%%%%%%%%%%%%%%%###%*==-===++===::::..:-:::::.::.::.:..+-:...::::::.::.:.::-::::..................::::.-:..........::............-----::
#%%%%%%%%%%%%%%%%%%%%%%%%%%%%%%%%%%%%###%+======++===:::::::-=::::::::.:.::.:=::..:::::::::::.:::::....................:..:.-...........::............:-----::
#%%%%%%%%%%%%%%%%%%%%%%%%%%%%%%%%%%%%##%+======**====::::::---:::::::..:::::=-.::::::::::::...:-::....................:::::-...........:-:...........::---::::
#%%%%%%%%%%%%%%%%%%%%%%%%%%%%%%%%%%###%+======%#+===::::::::=:::::::::::.::-+:::::::::::....::=.:......::...........::.:::-...........:-:............:---:::::
#%%%%%%%%%%%%%%%%%%%%%%%%%%%%%%%%%%##%+=====+##*===-:.::::-=:::::::::::::::-=::::::::::::::.:=:::...::::..........::::::::....::::::::--:......::::.::-:::::::
#%%%%%%%%%%%%%%%%%%%%%%%%%%%%%%%%%###*=====*##%===-:::::::--::::::.:::::::-+-:::::::::::::::=..:::.::::...::...:::::::::-..:::::::::::-:.....:::::.:::::::::::
#%%%%%%%%%%%%%%%%%%%%%%%%%%%%%%%%###*=====*##%+===-:::::--=:::::::::::::::-+-::::::::::::::=.:::::::::::::::::::::::::--..:::::::::::--:..::::::::::::::::::::
#%%%%%%%%%%%%%%%%%%%%%%%%%%%%%%%####=====*####===-::::---=:::::::::::::::-*=-::::::::-::.:=::::::::::::::::::::::::::--:::::::::::::--::::::::::::::::::::::::
#%%%%%%%%%%%%%%%%%%%%%%%%%%%%%%%###+====#%##%====-:::-----::::::::::::::::*--::::::::::::=-:::::::::::::::::::::::::=::::::::::::::-=:::::::::::::::::::::::::
#%%%%%%%%%%%%%%%%%%%%%%%%%%%%%%##%+====####%+====:::----=-:::::::-:::::::++--:::::::::::--:::::::::::::::.:::::::.:-::::::::::::::-=::::::::::::::::::::::::::
#%%%%%%%%%%%%%%%%%%%%%%%%%%%%###%+====##%#%#+==--:::-----::::::::::::::::#=---:::::-:::-=-:::::::.:::::::::::::::--::::::::::::::-=-::::::::::::::::::::::::::
#%%%%%%%%%%%%%%%%%%%%%%%%%%%###%*====######*===--.:-----::::::::-::::::::*=--::::--:::-==::::::::::-::::::::::::-=::::::::::::::-+=::::::::::::::::::::::::::=
#%%%%%%%%%%%%%%%%%%%%%%%%%%%%#%*=+==*######===---::-:--=::::::::::::::::+=----:::-::::=+:::::::::-:::::::::::::--::::::::::::::-++-:::::::::::::::::::::::::-=
#%%%%%%%%%%%%%%%%%%%%%%%%%%%###+===*#####%+===--::----=::::::::-::::::::#+---:::=-:::=+.::::::---::::::::::::-=:::::::::::::::-+*+::::::::::::::::::::::::::==
#%%%%%%%%%%%%%%%%%%%%%%%%%%###+===*#####%*=====-:------:::::::::::::::::#+==-::==-::=+::::::----::::::::::::-=::::::::-::::::-+#*-:::::::::::::::::::::::::==-
#%%%%%%%%%%%%%%%%%%%%%%%%%###*=+=#%#####%===-=-::----+-:::::::-::::::::-#==--:===:-==+:::::----::::::::::::==::::::::-::::::-**++-::::::::::::::::::::::::-=--
#%%%%%%%%%%%%%%%%%%%%%%%%%###===#%#####%+===---:----=:::::::::--:::::::=*==-:-==:-==+::::-----:::::::::::-==::::::::--:::::=**=++::::::::::::::::::::::::-=---
#%%%%%%%%%%%%%%%%%%%%%%%%%#%+==*###%%#%#==----------=::::::::::::::::::**-==-======+-:::-==--:::::::::::=+=::::::-:--::::-+#+==*-:::::::::::::::::::::::-=----
#%%%%%%%%%%%%%%%%%%%%%%%%#%*+=*%#%####%===---------=:-:::::::--::::::::*+-========*=::--===-::::::::-::-+:::::::----::::-**===*+::::::::::::::::::::::::=----+
#%%%%%%%%%%%%%%%%%%%%%%%###+=+#%#####%+=----------=---:::::::--::::::::*+-=======+=---=====::::::-:---==::::::--:==::::-#+====*-:::::::::::::::::::::::=----=+
#%%%%%%%%%%%%%%%%%%%%%%%#%==+#######%#==-----:----=-:-:::::::--:::::::-*+-=======++-======-::::-----==-:::::---==*::::=*=====++:::::::::::::::::::::::=----=*:
#%%%%%%%%%%%%%%%%%%%%%%##*++*%#%#%###+=----------=----:::::::::-::::::-**=======**======--:::---:--=+-:::----=+#*:::-++=====-*:::::-:::-:-::::::::--:-----=*=:
#%%%%%%%%%%%%%%%%%%%%%%#%+++%######%*=------:----=-:---::::::-::::::::=*+======**======-::-------==+:::-----+**+:::=*=====--+=:---::---:--:::::----------+#*-:
#%%%%%%%%%%%%%%%%%%%%%#%*+=%########=-------------:------:::--::::::::+*++=====+*======---------=++::-----=#*++-:-++====--::*::-::-------::::-----------+**=::
#%%%%%%%%%%%%%%%%%%%%##%+=#%#######+-------------=::--::-:::-=::::::::=*=+====*#======--------=+*=------=*+=++-:=+====--::.+-----------=-::--------=--=*+++-::
#%%%%%%%%%%%%%%%%%%%%##*++%#%%#####-------------=:::-::::-::-=:-::::::=*=+===+*=====+===--===+*+====-=++==+++--+===--:::::-=----------=-:---------=--+*==*-:::
#%%%%%%%%%%%%%%%%%%%%##*=#%#######--------------=-::----::-:-=::::::::+#++===**======------=**+-----++=+++++=++==--:::::..=----------=-----------=--++==+=::::
#%%%%%%%%%%%%%%%%%%%###++#######%=-------------=:::----:::::-=:::::-:-+#++==+*======----==***=---+***+++**+=+=--==-:::::.-=---------+=------------+*===++:::-+
#%%%%%%%%%%%%%%%%%%%%#*=*%#####%=--------------=:::----:::::-=:-:::-::=#+++=++=====----+*++*---++==--==-+====+++=+++-::::==--------*=------------*+===++-::===
#%%%%%%%%%%%%%%%%%%%##*+#%####%*=-------------=::::-:--:::-:--::::::::-*+++*#=====---+**+++--*+=======-++====--=----=*+.-+--------*-----------=+*==+==+---=+==
#%%%%%%%%%%%%%%%%%%%##*+#%#####=-----------=---::::-::-::-:::-::-:--::-+++++====---+#+++*+-#++++++++===--=--=----------=+-------=*=---------==*=====-+---+==--
#%%%%%%%%%%%%%%%%%%###++######+-----------==-=:::::+-:----:::--::::::::+++*#+==-=+#*+++*+*#*=======-==---=+*+==---------=------+*=---------+++==----=-:-*===--
#%%%%%%%%%%%%%%%%%%###++*####+------------=--::::::+:::--:-::-=::-:---:++++=--=***+*%@%#%%%%%%%%%%#+==-----=-=**=:-----+------=*=--------==+==-----==-+=-=----
#%%%%%%%%%%%%%%%%%####+++%##*---=--------==--::::::+-::::::::-=::::::::**+*+=+#+*#%%%%%%%%%%%%%%%%%%%%%#+=--:----+-:::-------*++--------=+=-----::=--+-----=++
#%%%%%%%%%%%%%%%%%%###+***%#=--=--:-----=+---::-:::+::::-::::----:----=*#**##%%%%%%%%@%@%%%%@@%%%######%%%##*=-:----::+----=*=+-------=+=----::::=-==-------::
#%%%%%%%%%%%%%%%%%%###+***#*-----:-----=+=---::::::+::::::::::=-::::---+%%%%%%%%%@%%#**+++=%@%%%%%#%##%%%%###*=--::::=----++=+-------+=---::...:=-==--::::::-=
#%%%%%%%%%%%%%%%%%%###*+***-----------=++---:-:::::+:::::--:::---::-----#*##*%%%%%*++++++++%@%%%%%%%%%%%%%%%%##=-::.=---=+===------+=---:.....:=+=--::.:.:----
#%%%%%%%%%%%%%%%%%#%%##+**----:-:----=+*----:-::::-+::::::::::-+----::-:#*#*=+%%%%#+++=++++%%%##%%%%#%+:*%%%%%*#*--:=-=+--==-----+=---:.....:-=--::.....::::-*
#%%%%%%%%%%%%%%%%%%%%##**=-=--------==*+---:::::::-+:::--:::::--:-------%%%*===%%%#+==-:::-%%*###%##%#+=#%%%%%%=+*-=+-:--==----==--:.......:==::..........:=*+
#%%%%%%%%%%%%%%%%%#%%#%#+-+---------=+*---::::::--==::::::-----=+:-----:-##%*===+%%*:......####***++#%%%%%#*##%=--=-::::-=---+--::........=-:............:===-
#%%%%%%%%%%%%%%%%%#%%###==:-------=+*++=--:::::::-===:::::--:--===----:-:%#%*=====*#:......-*#****++++%%%%#***%-.::::::---=:::::........................:::::.
#%%%%%%%%%%%%%%%%##%###+=--:-:----+*=+-=--:::::::-==#:::::::--::=+:------+%%*=====-+#::.....**#++=+=++*##*+++*=:......=--:..::............................:...
#%%%%%%%%%%%%%%%%%%###*--::-:----+%*++=--:::::::-===+:::::------==*:---:--%%%+=-=-::-+:......++++============+:.....:-:.......................................
#%%%%%%%%%%%%%%%%#####=-::::----*%#++=---:::::::-=====:::--------=+*:-----+%%+===-::::-=:....:=**+==--:---:-=-................................................
#%%%%%%%%%%%%#%%%####+--::::---*%##=+=--::::-:::-====*::::-------==++:----=%@%==-::.:::==:.:...:-++-:::-:--==:................................................
#%%%%%%%%%%%%%%%%###%--::::---*%#%+++=--::::-----+====-:::--------+++=----:=%@==-...:.::.:-:::::.:-===--=-::..................................................
#%%%%%%%%%%%%%%%###%+-::::---+%##%=++=---::------++==++::::--------=+*------*%%--::::::::::.::-=-=+++++=--:::.................................................
#%%%%%%%%%%%%%%####%=--::---=####*+=+---:-:--::--+++===-::::--=----=++*-----+%%*-:...::.::..::.::::...::...:..................................................
#%%%%%%%%%%%%%%####*---:---=#%##%+==+=--::-=--:--=====+=::::-------=+++#=--::#%%-:....::::.:...:.::.:..:.............................::::.:...................
#%%%%%#%%%%%%%#%###=-------#%###%=====--:--=-----==++=++=::::-==----=++*%*:--:#%#:::.:..::.:..::...:....:...........................::::::....................
#%%%%%%%#%%%%######-------*%####%=====----==-----=+++====-::::-+=----++*%%*-:-:#%=:.....::...:..:::.....:.....................:.....:::::::.....:.............
#%%%%%%%#%%%%%####+------+%######=====----==---:-=++=--==--::---=----=+*%%##+---#*:...::....:......................................::::::::::.................
*#%%%%%%%%%%%%%###=-----=######%*=======--==-::--=+++--==+-::::--=----=+*%++*#=--*+::.:......:....:................................:::::::::::................
#%%%%%%%%%%%%%%###=----=#######%*=======-==--::--=+++-:-==+-::::-*-----++%*+++**--*=:::...........................................::----:-:::::...............
#%%%%%%%%%%%%####*-----*#######%+=======-==---:---=+=::-=====::-:-#----=*@%#*+++*+-=+:::..................:.......................----------:::...............
#%%%%%%%%%%%%####+---==########%+=======-==--:-:--===-::====+=::::-+----*#%##%##*+**===:..:.......................................+-----:--:::................
#%%%%%%%%%%%%####+---=*%###%###%+==========----:--===-::-======::-:=+----+%%####%%%%%++=:....:...................................:+--:-:--:::.................
#%%%%%%%%%%%%####+--==#%##%#####*=========+-------===-::-=======-::-==---*%#########%%=--=+-.....................................::-:::::::...................
#%%%%%%%%%%%%####+-==+%#%#%######=========+=:------==-:::-======*::::-----*%%####%###@#-::.:.::...................................::::::......................
#%%%%%##%%%%%####*-+=+%%##%%%####=========+=-------==-:-::=======+-:::-=--=*#####%%##*%#:..::........................................:........................
#%%#%%#%#%%%%####*+=-*%%#%##%###%=========+=------:-==:::::========*:::----+*####@%###%%+:.:.:................................................................
#%%%%%%%#%%%######++-#%%%%%######*======++#+--------==::::::=======+*=--:--=+####%####%#%+:.......:...........................................................
#%%%%%%%%%%%%####%+==#%#%#%%%#####======++#*=-------==::--:::======+*++=:::-==###%####%##%=::.................................................................
#%%%%%%%%%%%%%####*==##%##%%###%#%=====+++#%=--------=-::::::-=====+*+***=---==#%%%#**%#*##+..:...............................................................
#%%%%%%%%%%%%######==**%##%%##%####======+#%*---------=::-::::-====+*+*****+---=*%%###%#*###*:.:..............................................................
#%%%%%%%%%%%%#%%##%++++############+=====+##%=---------:::-::::====++*********+--=*#**%#*##*##=....:...........................:==------=--:..................
*%%%%%%%%%%%%%#####%++**###%########+===++####---------=-:-:::::====+*********%#*+-=*##*****#*#*-..:.........................+***-::::::::-*++-:.:...........:
#%%%%%%%%%%%%%%##%###++**#%#########*====+*###+-----------:+::::-===+********%@#**#**+=*#***#####*::..........................::------=++===-::............:..
#%%%%%#%%%%%%%%%%%####*****%#########+====*###%---------:::==--::-==++*******%@*##*****#**######**%#-..................:....::::::::::::::::::::.....:....:..:
#%%%%%#%%%%%%%%#%%#####***+*#########%+==+*#####-----------:=-:-::-==********%%#****%####****###**%%%#=:..:.................:.::::::::::::::::::...:.:::.:::::
#%%%%%%%%%%%%##%%#######*****#########%===+#####*-----------:--::::-=+******#%#%***#****##***##*##*%%%%%#-:............:...::.::::::-::::::::::::...::::::::::
#%%%%%%%%%%%%%%%%#%######%*****########%+=+*#####*---------------:::-=+*****####%****####%%#########%%%%%%%*-:.::.......:...::.::::------:..::.:.:::::::::::::
#%%%%%%%%%%%##%%%%%%%######%#############*++######*-----------=------=+*****%####%#***####%%%##*#%%###%%%%%%%#=:::......::...::::::..::...::.:...:..:.:.::::--
#%%%%%%#%%%%%#%%%%%%%#%%##########%#######*++%######=----------=------=*****%##%@%%#**###%%@@@%###@@%%#%@@%%%%%%+-..::.::::::::::.::.:.::......:.:::::::::--==
#%%%%%%%%%%%%%#%##%#%%%####%###%%%##########**#######+----------=------=+***%@%%%%%%%***#%%%@@@%@%%@@@@%%%%%%%%@%%*=:::.:::.:::::::::......:..:::::::::----=+#
#%%%%%%%%%%%%%%%%%%#%%%%%%%%%##%%%%%###%%%%###########*+-=-------==-----=+**#%%%%%%%%%%**#%%@%@@@@@%%%%%%%%%%%%@@%#%#+:::::::::.:..:.:.:::.:::.::::::----+#%%#
*%%%%#%%%%%%%###%%%#%%%%%%#%%%%%%%###%%%##%%####%#%#####*+=-------==-----#**#%%%%%%%@@%%%#*%%@@%@%%@%@@%%%%%%%%@@%***#%*=::::::::::::.:::::..::::.:---=#%%####
*%%%%%%%%%%%%%%%%%%%%%%%%%%%%%%%%%####%%%%%%%%%##%########%--------==-----#*#%@@%%@%%%%@%%%%#%%@@@@@@@@@%@%%@%%%@@#+****%%+:::::::::::::.:::::::---=*%%######*
#%%%%##%%%%%%#%%%%%%%%%%%%%%%#####%##%%%#%###%%######%####%%--------+=-----+#%%%%%%%%%%%%%%@@@@@@@%%%%%%@@%%%%%%@@#++****##%%=:::::.::::::::::--=#%%########++
#%%%%#%%%%%%%#%%#%%#%#%%%%%%#%%##%%%%%%%%%%%%%%%#%#####%###@%=------=+=----=#%#%%%%%%%%%%%%%%@%%%%%@%@%@%%%%%%%%@%#=+++##*#*#%%#=-:::::::::-=+#%%%#######*++++
##%%%%#%%%%%%%%%%%%%%%%%%%%%%%##%#%%%%%%#%%%%#%%%%%######%%#*#+-------=-----=#%%%%%%%@%@@%%@%%@%%%%%%%%@%@%%%%%%@@%+++++*#######%%%+=---==+%%%#########*++++++
#%%#%%%%%%%%%%#%%%#%%%%%%%%%%%%%#%#%%%%%#%%%%%#%%%####%%%#####%+--==---=+----*%%%%%%%%@@@%%@%@@@@@%%@%%@%%%%%%%@@#%%%*+++**########%%%%%%%############++++++*#
#%%%%%%%#%%%%%#%%%%%%%%%%%%#%%%%%#%%%%%%%%%%%#%%%%##%%%#####%%%%*-==---===----#%%%%%%@%%%%@%%%@%%%%%%%@%%@%%%%%@@%%##%@@%#***######################****#%%%%%%
\end{lstlisting}
    \end{minipage}%
  }
\end{center}
\thispagestyle{empty}

\end{document}